\documentclass[lettersize,journal]{IEEEtran}
\usepackage{amsmath}
\usepackage{algorithmic}
\usepackage{algorithm}
\usepackage{array}
\usepackage[caption=false,font=normalsize,labelfont=sf,textfont=sf]{subfig}
\usepackage{textcomp}
\usepackage{stfloats}
\usepackage{url}
\usepackage{verbatim}
\usepackage{graphicx}
\usepackage{cite}
\usepackage{multirow}
\begin{document}

\title{No Place to Hide: Dual Deep Interaction Channel Network for Fake News Detection based on Data Augmentation}

\author{Biwei~Cao,
        Lulu~Hua,
        Jiuxin~Cao,
        Jie~Gui,~\IEEEmembership{Senior Member,~IEEE},
        Bo~Liu,~\IEEEmembership{Member,~IEEE},
        and~James~Tin-Yau~Kwok,~\IEEEmembership{Fellow,~IEEE}

\IEEEcompsocitemizethanks{\IEEEcompsocthanksitem Biwei Cao, Lulu Hua, Jiuxin Cao, and Jie Gui are with the School of Cyber Science and Engineering, Southeast University, Nanjing 211189, China, and also with the Key Laboratory of Computer Network and Information of Ministry of Education of China, Nanjing 211189, China.\protect\\
E-mail: \{caobiwei, hualulu, jx.cao, guijie\}@seu.edu.cn.
\IEEEcompsocthanksitem Bo~Liu is with the School of Computer Science and Engineering, Southeast University, Nanjing 211189, China.\protect\\
E-mail: bliu@seu.edu.cn.
\IEEEcompsocthanksitem James Tin-Yau Kwok is with the Department of Computer Science and Engineering, The Hong Kong University of Science and Technology, Hong Kong 999077, China.\protect\\
E-mail: jamesk@cse.ust.hk.
}

\thanks{Manuscript received April 19, XXXX; revised August 16, XXXX.}
\thanks{(Corresponding author: Jiuxin Cao, Jie Gui)}}
\markboth{Journal of \LaTeX\ Class Files,~Vol.~14, No.~8, August~2021}%
{Biwei \MakeLowercase{\textit{et al.}}: No Place to Hide: Dual Deep Interaction Channel Network for Fake News Detection based on Data Augmentation}

\IEEEpubid{0000--0000/00\$00.00~\copyright~2021 IEEE}

\maketitle

\begin{abstract}

Online Social Network (OSN) has become a hotbed of fake news due to the low cost of information dissemination. 
Although the existing methods have made many attempts in news content and propagation structure, the detection of fake news is still facing two challenges: one is how to mine the unique key features and evolution patterns, and the other is how to tackle the problem of small samples to build the high-performance model. Different from popular methods which take full advantage of the propagation topology structure, in this paper, we propose a novel framework for fake news detection from perspectives of semantic, emotion and data enhancement, which excavates the emotional evolution patterns of news participants during the propagation process, and a dual deep interaction channel network of semantic and emotion is designed to obtain a more comprehensive and fine-grained news representation with the consideration of comments. Meanwhile, the framework introduces a data enhancement module to obtain more labeled data with high quality based on confidence which further improves the performance of the classification model. Experiments show that the proposed approach outperforms the state-of-the-art methods.
\end{abstract}


\section{Introduction}
\IEEEPARstart{O}{nline} social media is a mixed blessing for news, which offers a platform for facilitating the dissemination of true news and fake news at the same time. In most cases, when breaking news emerges, fake news in online social media runs ahead of the truth, the proliferation of which seriously endangers government credibility and social stability. Therefore, it is urgent to improve the fake news detection ability in online social media.


Some researches focus on the propagation structure to do fake news detection and our group also has made contributions to this field \cite{suntkde, liutnnl}.
Some scholars have made many attempts  \cite{zhou2018fake} on news content as well. These works have tried to detect fake news automatically from the perspectives of emotional polarity  \cite{castillo2011information}, linguistic style  \cite{popat2017assessing}, and chronological features  \cite{ma2016detecting} on news-related comment semantics.


Besides, existing works have been considering emotion features often as parts of article styles (e.g. the proportion of emotional words  \cite{2019Emotion}, the emotion consistency between news and comments  \cite{zhang2021mining}) to obtain simple features of emotion signals, since fake news contents are always designed to stir emotions of the crowd who read the pieces and so the sentiment factors are also of great importance for fake news detection. 

 However, these methods destroy the integrity of the individual comment and ignore the emotional change from comment to comment. We observe that emotion evolutionary pattern is essential to further improve the accuracy of fake news detection. The modelling of emotion evolutionary pattern has the capability of indicating mood swings of the commenters and is used by human experts in fake news detection. 
True and false news have different emotional evolution patterns, which is essential for identifying news authenticity. 
To be more specific, the crowd emotion of true news fluctuates less over time and the emotion polarity tends to be consistent overall, while the emotion of fake news fluctuates relatively stronger and the emotion polarity is more inconsistent. These differentiated time-series emotional characteristics are instructive for this study.



In addition, it is generally difficult for us to get enough fake news samples to train a useful classification model, so the scarcity of labeled data is another challenge for fake news detection tasks. Therefore, designing an appropriate method of making full use of the limited labeled data, to further improve the performance of the data-driven fake news detection task, is also one of the focuses of this study.

\IEEEpubidadjcol  In conclusion, this paper aims to improve the accuracy and efficiency of fake news detection from two perspectives: extraction of key features of semantics and emotion, and enhancement of limited labeled data. In detail, we investigate into the following three research questions.

\begin{itemize}
    \item RQ1: Are the chronological properties of news commenter emotion signals useful in fake news detection tasks?
    \item RQ2: Can multi-dimensional features of semantics and emotion be integrated more effectively?
    \item RQ3: How to make good use of the limited samples to enhance the dataset to get a high-quality detection performance?
\end{itemize}

Based on the above, the main contributions of our work can be summarised as follows.

Firstly, we are the first to model the news commenters' emotion evolution from a chronological perspective and experimentally validate its effectiveness for fake news detection (RQ1), which is distinguished from existing studies that only model semantics and communication structure chronologically, or works that treat all comments as one long sentence to model the comment emotion.

Secondly, we design a dual deep interaction channel module to fully fuse the multi-dimensional features of news and comments. In each channel, the interactive co-attention is used to deeply align the features of news and comments, and the chronological emotion is embedded by Bi-GRU temporal model,  and finally, the multi-dimensional features from the two channels are fused together by concatenation (RQ2).

Finally, we enhance the labeled examples based on back translation and other data augmentation techniques, and the pseudo-label technology is utilized to screen the noise labels based on confidence, so as to obtain more high-quality labeled data, further improving the model classification performance (RQ3).

\section{Related Work}

\subsection{Fake News Detection}
Fake news detection research works can be mainly divided into two categories: content-based and propagation-based.

\subsubsection{Content-based}
Content-based fake news detection works have modeled various dimensions of news content style  \cite{popat2017assessing}, comment emotion  \cite{2011Information}  \cite{2019Leveraging}, user portraits  \cite{2020Fake}, etc., which is conducive to early detection of fake news  \cite{yuan2020early}  \cite{2018Early}. However, they only consider the content semantic features of the comments themselves, ignoring the important chronological emotion features of user comments, and therefore lacking the dynamic global evolution features of crowd emotion in the comment information.

\subsubsection{Propagation-based}
Propagation-based methods study the propagation structure features of fake news distinguished from real news by information cascade or propagation graph. The performance of this type of detection approaches depends on the integrity of the propagation chain structure  \cite{2013Epidemiological}  \cite{2017Detecting}  \cite{2015False}  \cite{2016Leveraging}. 
In the early stage of fake news propagation, such methods often perform poorly due to incomplete information on propagation structure. 

\subsection{Data Enhancement on Fake News Detection}

The small benchmark datasets for fake news detection (e.g., the RumorEval-19 dataset only has 446 labeled examples with 327 training labels) severely limit the improvement on the fake news detection task performance. Related works on labeled data enhancement for fake news detection tasks is not too much. Some researchers have tried to enhance the examples based on weakly supervised learning, which use user reports as weakly supervised signals, and combine with reinforcement learning \cite{2020Weak}. However, the method of utilizing user reports has a poor generalization since the datasets are not publicly available and have their own particularity such as Tencent WeChat user data. 
In this paper, we enrich the data by simple and common data augmentation methods used in NLP \cite{2021Data}, and select highly trusted pseudo-labeled examples \cite{2015Self} \cite{2021In} based on confidence, which improve the augmentation effectiveness and keep the purity of data at the same time.

\section{Methodology}
\subsection{Problem Definition}
\subsubsection{Input} Given the formatted data set $S$, and $m$ pieces of data, we have 

\begin{equation}\nonumber
S=\left\{\left(n_{1}, c_{1}, t_{1}, l_{1}\right),\left(n_{2}, c_{2}, t_{2}, l_{2}\right) \ldots,\left(n_{m}, c_{m}, t_{m}, l_{m}\right)\right\},
\end{equation}
where$\left(n_{k}, c_{k}, t_{k}, l_{k}\right) \in S$ is the primary field for a specific piece of data in the dataset, $n_{k}$ denotes the $k$-th piece of news text, $c_{k}$ denotes the set of comments corresponding to the $k$-th news' comments, $t_{k}$ denotes the timestamp corresponding to the comment under the $k$-th news item, $l_k \in \{0,1\}$ denotes the label of the $k$-th news item, and one piece of news corresponds to one label.
\subsubsection{Output}A prediction label for the news, either 1 (fake news) or 0 (true news).

\subsection{Overall Framework}

\begin{figure*}[t]
\centering
\includegraphics[width=2\columnwidth]{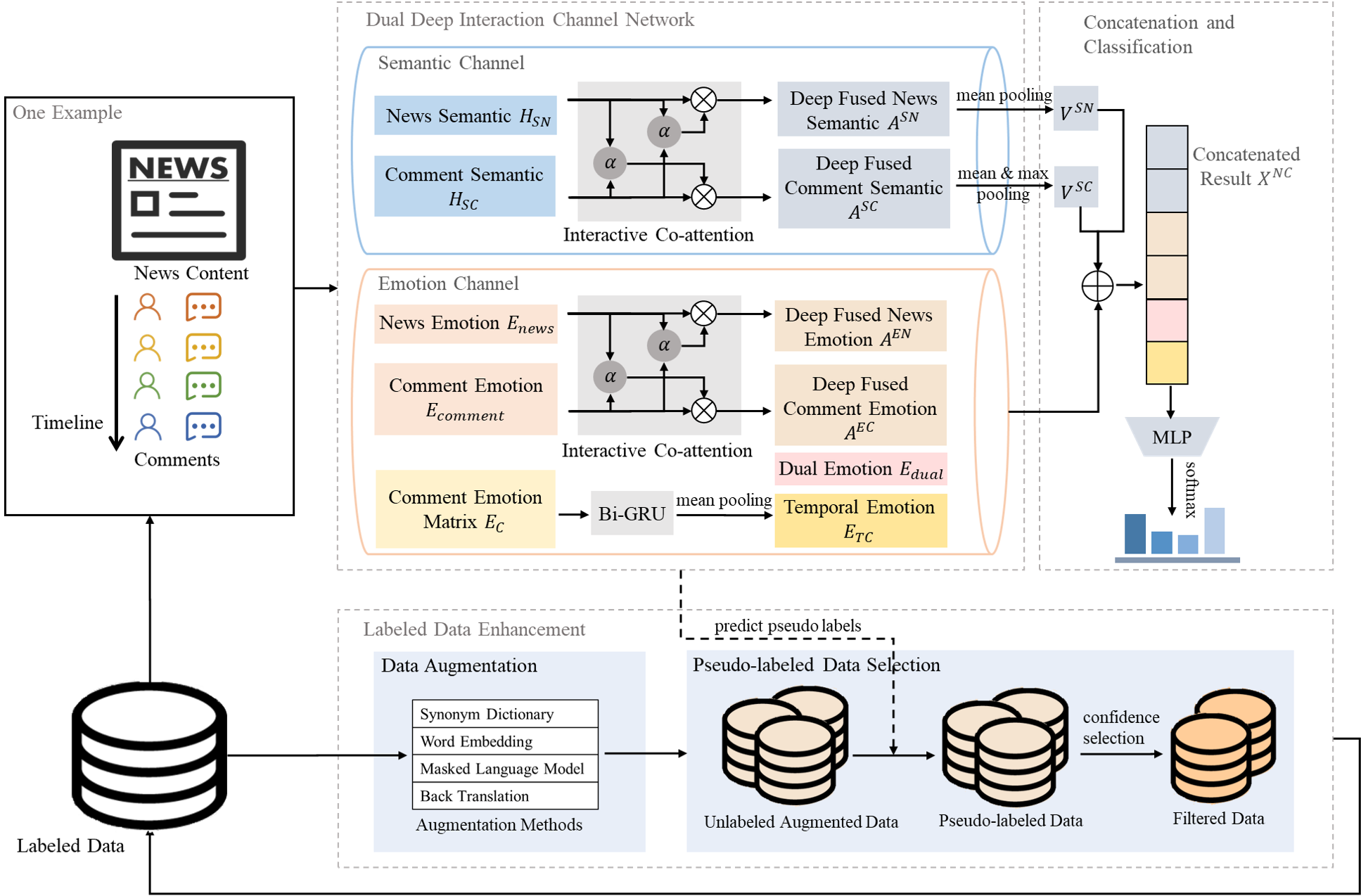} 
\caption{The overall framework. This framework consists of three modules. The dual deep interaction channel network module models the chronological emotion evolution of comments and extracts the fine-grained emotional and semantic features of news and comments. The concatenation and classification module concatenates the features obtained from the dual deep interaction channel network and finally gets the classification result of the given new piece by inputting the concatenated result into the softmax function. The labeled data enhancement module augments the given dataset and labels the expanded unlabeled data by the classifier achieved from the last module. Then expanded data with high confidence is selected as the supplementary of the given dataset.}.
\label{fig2}
\end{figure*}

In this paper, we propose a fake news detection framework called \underline{D}ual Deep \underline{I}nteraction Channel Network based on \underline{D}ata \underline{A}ugmentation framework (DIDA), shown in Figure~\ref{fig2}.

Our framework consists of three modules. The first module is the dual deep interaction channel network module. This module firstly performs dynamic chronological modeling of comment emotion, and extracts the fine-grained emotional and semantic features corresponding to news and its comments respectively. The dual channels mine the deep association (consistency or conflict) of different features between the news and its comments by interactive co-attention calculation. 

The second module is the concatenation and classification module. This module concatenates the emotional and semantic features first. Then the concatenated vector is sent into the softmax function to achieve the classification result of the given news piece.

The last module is the labeled data enhancement module. This module augments the given labeled dataset by using the substitution and back translation methods, then predicts the expanded unlabeled data by the detection classifier trained in the second module based on the given dataset, and finally selects the high confidence examples as a supplement to the existing labeled dataset, making full use of the limited labeled data and providing a solution to the model performance problem caused by the limited labeled examples in the fake news detection task.


\subsection{Dual Deep Interaction Channel Module}

The dual deep interaction channel module consists of three parts: 1) semantic channel: extract the semantic features of the news and its comments; 2) emotion channel: mine the dynamic emotion evolution pattern of comments and extract the emotion features of the news and its comments; 3) interactive co-attention: applied in two channels and reassign weights to the obtained features by the co-attention mechanism, to achieve a deeper feature fusion.

\subsubsection{Semantic channel} News and its corresponding comments are both short texts in essence. Therefore, we can use a similar method for obtaining news and its comment semantic representations. For the $i$-th text statement $T_{i}$, the word embedding vector is formed by finding the vector representation corresponding to the current word in the word vector list to achieve the mapping of text symbols in the vector space $F_{g} \in R^{n \times d_{g}}$, where $n$ is the number of words contained in the text $T_{i}$, and $d_{g}$ is the dimensionality of the word vector \cite{2014Glove}. The process can be formulated as
\begin{equation}
F_{g}=\operatorname{GloVe}\left(T_{i}\right).
\end{equation}

To improve the training efficiency, this study adopts BiGRU \cite{2014Empirical} to model the sequence relations of the hidden semantics of text word embedding, to obtain the hidden semantic features incorporating contextual information.

Denoting the text $T_{i}$, the context vector $H_{S}$ and the semantic features $V_{S}$, we have 
\begin{equation}
H_{S}=\operatorname{BiGRU}\left(F_{g}\right).
\end{equation}

Both news and comments share the same feature extraction process. Let the $H_{SN}$ denote the semantic features corresponding to the news text and $H_{SC}$ denote the semantic features corresponding to the comments under the news piece.


So far, the semantic features of news $H_{SN}$ and comments $H_{SC}$ are obtained.

Then we do the interactive co-attention of the $H_{SN}$ and $H_{SC}$ to achieve the deep fused news semantic feature $A^{SN}$ and comment semantic feature $A^{SC}$. The detailed information is mentioned in ``Interactive co-attention".

\subsubsection{Emotion channel} 
\begin{itemize}
\item Fine-grained emotion feature extraction: 
this part aims to extract fine-grained emotion-related embedding representations. Specifically, we refer to Zhang et al.'s approach  \cite{zhang2021mining} from emotion category $E_{T}^{\text {cate}}$, emotion lexicon $E_{T}^{\text{lex}}$, emotion intensity $E_{T}^{\text{int}}$, emotion score $E_{T}^{\text{score}}$ and auxiliary feature $E_{T}^{\text {aux}}$. These five dimensions calculate the emotion features corresponding to the text $T$. 
    
    



Assuming the emotion feature corresponding to the text T is $E_T$, we have
\begin{equation} \label{eq55}
E_{T}=E_{T}^{\text {cate }} \bigoplus E_{T}^{\text {lex }} \bigoplus E_{T}^{\text {int }} \bigoplus E_{T}^{\text {score }} \bigoplus E_{T}^{\text {aux }},
\end{equation}
where the symbol $\bigoplus$ represents the concatenation operation, $E_{T}^{\text {cate }} \in R^{d_{cate}}$, $E_{T}^{\text {lex }} \in R^{d_{lex}}$, $E_{T}^{\text {int }} \in R^{d_{int}}$, $E_{T}^{\text {score }} \in R^{d_{score}}$, $E_{T}^{\text {aux }} \in R^{d_{aux}}$, $E_{T} \in R^{d}$ and $d = d_{cate} + d_{lex} + d_{int}+ d_{score} + d_{aux}$.

According to Equation \ref{eq55}, we can get the emotion characteristics corresponding to a news piece $E_{news} = E_T$.

Similarly, let the set of comments related to the news piece consist of $M$ independent texts $T$, i.e., comments $C=[T_1,T_2,…,T_M]$. The comment emotion matrix is $ E_{C}=E_{T_{1}} \bigoplus E_{T_{2}} \bigoplus \ldots \bigoplus E_{T_{M}} \in R^{M \times d} $, where the symbol $\bigoplus$ represents the concatenation operation.

With the aim to align with the dimensionality of the news emotion feature, this study performs the mean-pooling and max-pooling operations on the obtained comment emotion matrix and concatenates them together. The global emotion vector of all comments related to one news piece $E_{comment}$ can be formulated as 
\begin{equation}
E_{C}{ }^{\text {mean }}=\operatorname{MeanPooling}\left(E_{C}\right) \in R^{d},
\end{equation}
\begin{equation}
E_{C}{ }^{\max }=\operatorname{MaxPooling}\left(E_{C}\right) \in R^{d},
\end{equation}
\begin{equation}
E_{\text {comment }}=E_{C}^{\text {mean }} \oplus E_{C}^{\max} \in R^{2 d},
\end{equation}
where the symbol $\bigoplus$ represents the concatenation operation.

Then we do the interactive co-attention of the $E_{news}$ and $E_{comment}$ to achieve the deep fused news emotion feature $A^{EN}$ and the deep fused comment emotion feature $A^{EC}$ . The detailed information is mentioned in the ``Interactive co-attention".

\begin{figure*}[t] \label{fig3}
\centering
\includegraphics[width=1.2\columnwidth]{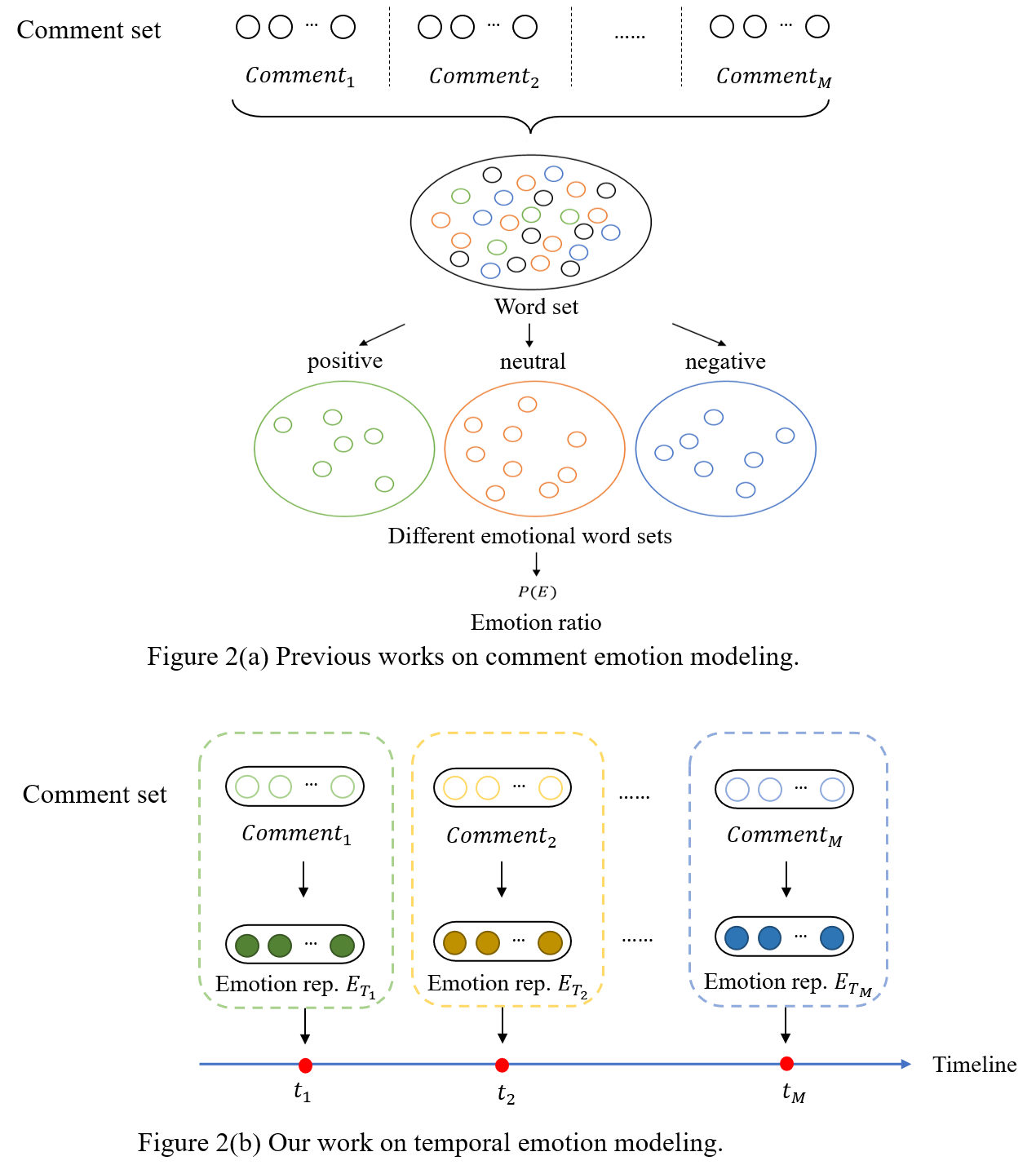} %
\caption{Innovation of temporal emotion modeling in our DIDA framework. The top is the previous works on comment emotion modeling and the bottom is our work on temporal emotion modeling. Different from the previous works, our temporal emotion feature preserves the integrity of the individual comment as an independent unit of emotional expression.}.
\label{fig3}
\end{figure*}

\item{Dual emotion feature extraction}: referring to the method of Zhang et al.’s approach \cite{zhang2021mining}, we calculate the emotion gap between the news and comments based on the emotion features above. The emotion gap is achieved from the concatenation of the two subtractions. One is the subtraction between the news emotion $E_{news}$ and comment mean pooling emotion $E_{C}^{mean}$ and the other is the subtraction between the news emotion $E_{news}$ and comment max pooling emotion $E_{C}^{mean}$. This process can be written as

\begin{equation} \label{eq3.16}
E_{gap}= (E_{news} - E_C^{mean})\oplus (E_{news} - E_C^{max}) \in R^{2d}.
\end{equation}

The dual emotion feature can be obtained from the concatenation of the news emotion, comment emotion and the gap emotion:
\begin{equation} \label{eq3.17}
E_{dual}= E_{news}\oplus E_{comment} \oplus E_{gap} \in R^{2d}.
\end{equation}

\item{Temporal emotion feature extraction}: 
some studies  \cite{2020Adaptive}  have attempted to consider the emotion as one part of the semantic, which collect the set of comments and concatenate the comments into one long sentence. However, in fact, no contextual relationship logically exists among the emotion words of different comments. This method ignores the integrity of the comment as an independent emotion expression unit. Figure~\ref{fig3} visualizes the difference between previous studies and ours. Assuming $M$ comments are in the set of comments for one news piece, $Comment_i$ denotes the $i$-th comment representation and $E_{T_i}$ denotes a sentence-level emotion representation of a single comment from Equation \ref{eq55}. 
By our method, a complete comment emotion representation is achieved by considering each comment as an individual.
This preservation of the completeness of sentence-level emotion vectors as independent emotion units with timestamp information is conducive to better representing the evolution pattern of news-related commenters' emotion over time. 

For the implementation, we use BiGRU model to compute the emotion vectors of comments in chronological order, and take the mean pooling of the BiGRU hidden vector from all comment emotions  as the global evolution feature of commenters' emotions under the news $E_{TC}$. 

This process can be formalized as 
\begin{equation} \label{eq9}
E_{TC}=\text { MeanPooling }\left(\operatorname{BiGRU}\left(E_{C}\right)\right) \in R^{d_{h}},
\end{equation}
where $E_C$ is the emotion matrix of all the comments and $d_h$ is the hidden layer dimensionality.

Therefore, the temporal emotional characteristics of the commenters are obtained.
\end{itemize}

\begin{table*}
\centering
\renewcommand\arraystretch{1.2}
\caption{Parameter settings on different datasets. }
    \begin{tabular}{c c c}
    \hline
    Parameters & RumourEval-19 & Weibo-16 \\
     \hline
    text maximum length & 50 & 100  \\
   
    comments maximum number & 345 & 100 \\
    learning rate & $1e^{-3}$ & $1e^{-3}$ \\
    GloVe dimensionality  & 200 & 300\\
    epoch  & 50 & 50\\
    batch size & 32 &32 \\
    L2  & 0.01 & 0.01 \\
    \hline

    \end{tabular}
    \label{tableparameter}
\end{table*}

\begin{table*}
\centering
\renewcommand\arraystretch{1.2}
\caption{Comparison experiment and ablation test results. }
    \begin{tabular}{c c c c c c}
    \hline
    \multicolumn{2}{c}{\multirow{2}{*}{Models}} & \multicolumn{2}{c}{RumourEval-19} & \multicolumn{2}{c}{Weibo-16} \\
    \cline{3-6}
    \multicolumn{2}{c}{~} & macro F1 & RMSE & macro F1 & accuracy \\
    \hline
    \multirow{5}{*}{Baselines} & EmoRatio & 27.5 & 82.3 & 79.4 & 81.0 \\
    ~ & EmoCred & 31.1 & 79.7 & 76.6 & 77.8 \\
    ~ & NileTMRG & 30.9 & 77.0 & - & - \\
    ~ & HSA\_BLSTM & - & - & 84.9 & 85.5 \\
    ~ & \textit{DualEmotion} & 34.2 & 75.6 & 90.8 & 91.3 \\
    \hline
    \multirow{3}{*}{Variants} & DIDA-T&34.7&75.7&91.8&91.8\\
    ~ &DIDA-D& 37.0& 74.6& 93.3& 93.3 \\
    ~ & DIDA-A&37.2&73.6&92.5&92.5 \\
    \hline
    \multirow{2}{*}{Ours} & \textbf{DIDA} & \textbf{46.4} & \textbf{69.1} & \textbf{94.8} & \textbf{94.8} \\
    ~ & \textbf{Impv.} & +12.2\% & +6.5\% & +4\% & +3.5\% \\
    \hline

    \end{tabular}
    \label{tab1}
\end{table*}

\begin{figure*}[t]
\centering
\includegraphics[width=2\columnwidth]{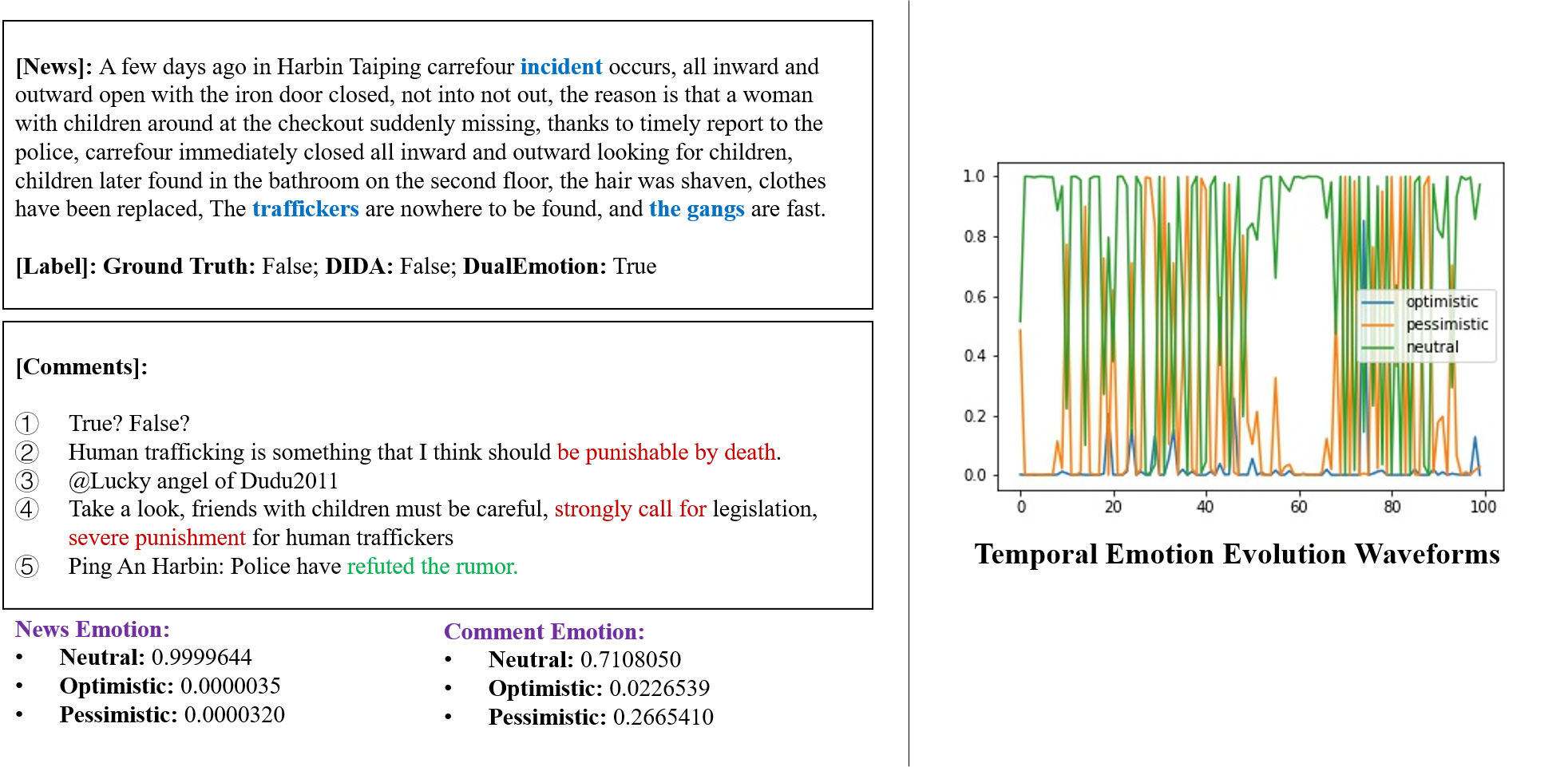} 
\caption{Fake news case study 2. The left are the news and comments contents with the emotion prediction scores. For the convenience of the display, the first 5 items in the set of comments are chosen for display. The right is the line chart of the temporal emotion evolution. }.
\label{fig6}
\end{figure*}

\begin{figure*}[t]
\centering
\includegraphics[width=2\columnwidth]{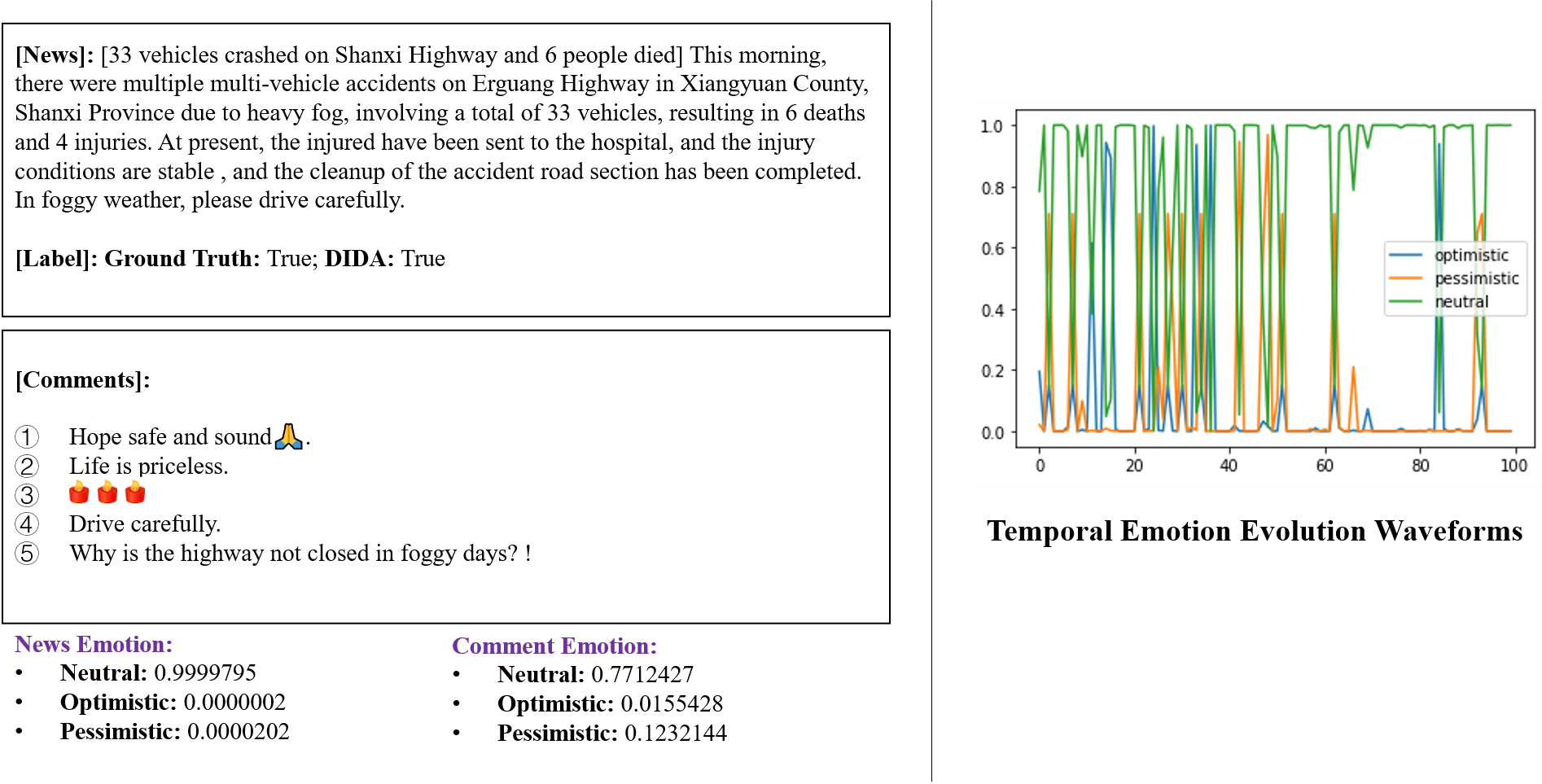} 
\caption{Fake news case study 2. The left are the news and comments contents with the emotion prediction scores. For the convenience of the display, the first 5 items in the set of comments are chosen for display. The right is the line chart of the temporal emotion evolution. }.
\label{fig7}
\end{figure*}



\subsubsection{Interactive co-attention} Most existing studies have used simple addition operations directly when obtaining multi-dimensional feature representations, failing to capture deep semantic connections and emotional associations between news and comments. 

To address the above problems, this study designs an interactive co-attention method for two channels to respectively capture the deep association information between news and comments at the emotional and semantic levels.

The process of computing  for the co-attention \cite{2014Neural} \cite{2016Hierarchical} of semantic channels can be expressed as

\begin{equation}\label{eq12}
\varphi\left(h_{i}^{S N}, H_{S C}\right)=\tanh \left(W_{S} h_{i}^{S N} H_{S C}+b_{S}\right),
\end{equation}
\begin{equation} \label{eq13}
\alpha_{{i}}=\frac{\exp \left(\varphi\left(h_{i}^{S N}, H_{S C}\right)\right)}{\sum_{j=1}^{m} \exp \left(\varphi\left(h_{j}^{S N}, H_{S C}\right)\right)},
\end{equation}
\begin{equation} \label{eq14}
\boldsymbol{A}^{S N}=\sum_{i=1}^{m} \alpha_{i} h_{i}^{S N},
\end{equation}
where the $h_i^{SN}$ or $h_j^{SN}$ denotes the $i$-th or the $j$-th hidden state of the news semantic feature $H_{SN}$. $H_{SC}$ is the semantic vector of the comments. $\varphi(\cdot)$ denotes the scoring function calculated by dot product instead of the simple addition operation. $W_S$ and $b_S$ are trainable parameters. $A^{SN}$ is the news semantic feature after the deep fusion by reassigning the weights of the association between the news semantic and comments semantic. Similarly, we can also achieve the comment semantic features  $A^{SC}$,  news emotion $A^{EN}$ and comment emotion $A^{EC}$ after deep fusion in the same way, shown in Equation \ref{eq12} to \ref{eq14}.

\subsection{Concatenation and Classification} This module first unifies dimensions of features obtained from the dual deep interaction channel module. We do the mean pooling operation on the deep fused news semantic features as 


\begin{equation}\label{eq3}
V_{SN}=\text { MeanPooling }\left(A^{SN}\right).
\end{equation} 

Since a news piece often corresponds to several comments, here we represent the number of comments to be $M$. A further operation max pooling is needed which is formulated in Equation \ref{eq4}.

\begin{equation}\label{eq4}
V_{SC}=\text{MaxPooling}\left(\text { MeanPooling }\left(A^{SC}\right)\right).
\end{equation}


Then, we do the concatenation of all the features and input the concatenated results into the softmax function to obtain the predicted probability distribution $P$ of the given news, as shown in Equation \ref{eq15} and Equation \ref{eq16}:

\begin{equation} \label{eq15}
X^{NC}=V^{S N} \oplus V^{SC}  \oplus A^{E N}\oplus A^{EC} \oplus E_{TC} \oplus E_{\text {dual }},
\end{equation}
\begin{equation} \label{eq16}
P=\operatorname{softmax}\left(W_{X} X^{N C}+b_{X}\right),
\end{equation}
where the $E_{TC}$ is the temporal emotional feature of comments. $W_X$ and $b_X$ are the parameters to be learned.

In this study, cross entropy loss is used as the loss function of the model. We utilize the value of ground truth label $y_r$ to minimize the cross entropy error of single training instance, as shown in Equation \ref{eq17}:
\begin{equation} \label{eq17}
\textit{\text { Loss }}=-\sum\left[y_{r} \log \left(y_{p}\right)+\left(1-y_{r}\right) \log \left(1-y_{p}\right)\right],
\end{equation}
where $y_p$ is the prediction label probability.



\subsection{Labeled Data Enhancement Module}

The labeled data enhancement module is made up of two parts: data augmentation and pseudo-labeled data selection.

\subsubsection{Data augmentation} The data augmentation part adopts simple and effective methods of data generation in NLP - the substitution method and back-translation method. The substitution method replaces a part of the original text without changing the semantics of the sentence. In this paper, three substitution strategies are adopted: synonymy dictionary based, word embedding based and masked language model based. The back-translation method is to generate more text variants by back-translating.

\subsubsection{Pseudo-labeled data selection} 

Pseudo-labelling technique  \cite{2015Self} is able to learn from both unlabeled and labeled data, which uses the class with the highest prediction probability as the labeled result for the current training instance. It is a special form of entropy minimization in semi-supervised learning and is a viable method for data augmentation using unlabeled data.

The label selection part filters to obtain trusted pseudo-labeled examples as trusted data supplement \cite{2021In} based on a confidence threshold:

\begin{equation} \label{eq18}
    \tilde{y}^{(i)}=1\left[p^{(i)} \geq \gamma\right].
\end{equation}
In Equation \ref{eq18}, $\tilde{y}^{(i)}$ denotes the pseudo-labeled data which is labeled by the fake news detection classifier after being augmented. $p^{(i)}$ is the probability of the classification model's output, and $ \gamma $ denotes the threshold value. 

After introducing the pseudo-label of negative examples, let $g^{(i)}$ denote whether the pseudo-label of this example  will be used in the training model or not. The process can be formulated as below:

\begin{equation}
    g^{(i)}=1\left[p^{(i)} \geq \tau_p \right] + 1\left[p^{(i)} \geq \tau_n \right],
\end{equation}
where $\tau_p$ and $\tau_n$ denote the selection thresholds of pseudo-labels for positive and negative cases respectively. Both of them are parameters that can be learned.

Then, we can get the loss function of our fake news detection model integrated with pseudo-labels: 
\begin{equation} 
\begin{split}
Loss(\tilde{y}^{(i)}, \hat{y}^{(i)},g^{(i)} )=\\-\frac{1}{s^{(i)}}\sum_{c=1}^{s^{(i)}}g_c^{(i)}  \left[ \tilde{y}^{(i)}_c \log(\hat{y}^{(i)}_c) + (1-\tilde{y}^{(i)}_c)\log(1-\hat{y}^{(i)}_c)\right],
\end{split}
\end{equation}
where $s^{(i)}$ is the number of examples with pseudo-labels and $\hat{y}^{(i)}$ is the original probability of the enhanced data by the classification model.

\section{Experiments}
\subsection{Experimental Setup}
\subsubsection{Datasets}
In this paper, we evaluate the performance of DIDA using the benchmark datasets RumourEval-19  \cite{2019SemEval} and Weibo-16 \cite{2016Detecting}. This study keeps the consistency with the division of the SemEval-2019 competition. The RumourEval-19 dataset is divided into a training set, a test set and a validation set in the ratio of 7:2:1 and the Weibo-16 dataset is divided into a training set, a test set and a validation set in the ratio of 6:2:2.
\subsubsection{Parameters}
In this paper, the training parameters of our proposed model on different datasets are listed in Table~\ref{tableparameter}.

\subsubsection{Experimental environment}
The main parameters of the server computing environment are as follows:
\begin{itemize}
    \item CPU: Intel(R) Xeon(R) Gold 6132 CPU @ 2.60GHz 14C28T × 2.
    \item Memory: 32GB DDR4 2666MHz ECC × 16.
    \item GPU: NVIDIA Geforce RTX2080, 8GB GDDR6 × 8.
    \item Operating system: CentOS Linux 7 (Core) / Linux 3.10.0.
    \item Software dependencies: Python 3.9.12 / TensorFlow 2.8.0 / Keras 2.8.0.
\end{itemize}

\subsubsection{Evaluation metrics}
On the dataset RumourEval-19, this study uses the official evaluation metrics \cite{2018RumourEval}, macro F1 score and $RMSE$ (root mean squared error). 
With the consideration of data imbalance, on the Weibo-16 dataset, accuracy and macro F1 metrics, are chosen to be the evaluation metrics, same as previous works \cite{2018Rumor} \cite{zhang2021mining}. We choose the five-folded cross-validation and report the average performance.

\subsubsection{Baselines}
This study is compared with state-of-the-art models for fake news detection tasks in recent years, including DualEmotion \cite{zhang2021mining}, EmoRatio \cite{2019Sentiment}, EmoCred \cite{2019Leveraging}, NileTMRG \cite{2017NileTMRG}, and HSA\_BLSTM \cite{2018Rumor}. 
\begin{itemize}
    \item DualEmotion: From WWW2021, it is a method to detect fake news by mining the emotional information related to news. By mining the dual emotional characteristics of Publisher Emotion and Social Emotion of news, the fake news detection task is modeled and analyzed.
    \item EmoRatio: It is an algorithm to represent the emotional features related to news \cite{2019Sentiment}. In this method, the ratio of negative words to positive words is calculated to represent the emotional characteristics related to news. In this study, BiGRU is used as the basic classifier for EmoRatio features for comparison fairness.
    \item EmoCred: It is a method to model false news from the perspective of emotion. Specifically, this method utilizes the emotion vocabulary and emotion intensity features of content text. These features are calculated based on the frequency of lexical symbols. BiGRU is used as the basic classifier for EmoRatio features for comparison fairness.
    \item NileTMRG: For the RumoureVAL-19 dataset, this study uses a model officially implemented by the competition organization, NileTMRG, which outperforms the vast majority of the listed contestants' models. The model is a linear SVM, which uses text feature, social feature and comment stance feature. In the experiment, we keep all the hyperparameters of the original model.
    \item HSA\_BLSTM: For the Chinese dataset Weibo-16, this study implements the baseline model HSA\_BLSTM, which is widely used in dataset Weibo-16. This model proposes a hierarchical neural network of attention, which simultaneously utilizes the relevant content of news and its comments. In the experiment, we follow the hyperparameter settings of the original model.
\end{itemize}

\subsubsection{Comparative experiment} Table~\ref{tab1} shows the performance comparison between the existing research works and DIDA, the framework proposed in this paper, on two datasets. In this table, the best results of evaluation metrics are in bold. On the RumourEval-19 dataset, DIDA has outperformed the SOTA model in both the $F1$ metric and the $RMSE$ metric, with the $F1$ value improved by nearly 12.2 percentages over DualEmotion and the $RSME$ performance improved by 6.5 percentage points. Here, the lower $RSME$ represents the better performance. The $F1$ value is increased by nearly 4 percentage points compared to DualEmotion, while the accuracy metric climbed by 3.5 percentage points. The comparative experiments validate the effectiveness of DIDA.

\subsubsection{Ablation test} In this paper, we focus on three variants of DIDA, including DIDA-T (the only difference from the baseline DualEmotion is the addition of chronological emotion features), DIDA-D (the addition of the interactive co-attention mechanism compared to DIDA-T) and DIDA-A (the addition of expanded data compared to DIDA-D, but without label selection process compared to DIDA).

As can be seen from Table~\ref{tab1}, DIDA-D has improved in $F1$ metrics compared to DIDA-T on both datasets, which indicates the effectiveness of the dual deep interaction channel network proposed in this study. 
While on the basis of deep fusion, the $F1$ metric of DIDA-A decreases compared to DIDA-D, which indicates that the training with direct NLP-based augmented data is less optimized for the learning process of the model. This even leads to a decrease in the learning performance of the model due to the introduction of noise, since the credibility of the labels of the augmented data cannot be guaranteed. 
By introducing the pseudo-labeling technique with confidence selection based on the data expansion of DIDA-A, it can be found that the final framework DIDA model proposed in this paper has increased on both $F1$ value and accuracy, which fully proves the effectiveness of the pseudo-labeling selection mechanism proposed in this study.

\subsubsection{Case study}
In this section, we present two real cases of the DIDA framework from the Weibo-16 dataset, to show the effectiveness of our model. Figure~\ref{fig6} is an example with ground truth label `` False" that our proposed model DIDA classifies correctly while SOTA model DualEmotion classifies incorrectly. Figure~\ref{fig7} is an example with ground truth label ``True" and our model DIDA classifies correctly. These two cases prove that 
the proposed model DIDA has the superiority to identify fake news compared to the SOTA model while maintaining the ability to identify real news. For the convenience of display, the first 5 items in the set of comments are chosen for display.

Figure~\ref{fig6} shows a case of fake news. From this figure, it can be seen that the overall emotion polarity of news is mainly neutral, while the overall emotion of comments is also neutral, which is also relatively interspersed with pessimistic emotion. In such cases, SOTA method DualEmotion fails to distinguish the truthfulness of news since the consistency and inconsistency correlation between news emotion and comment emotion are both obscure. In contrast, the DIDA method proposed in this study can correctly classify this case. From the chronological waveform diagram in Figure~\ref{fig6}, it can be seen that the comment emotion of this news changes relatively drastically, shown by the high frequency of steeping up and steeping down of the folded graph. Figure~\ref{fig6} well explains the superiority of DIDA in identifying fake news - when the emotion correlation between news and comments is not obvious, the temporal emotion feature designed in this study can capture the unique pattern of commenters' emotion evolution over time better in fake news.

Figure~\ref{fig7} shows a case of true news. Same as Figure~\ref{fig6}, the overall emotion polarity of news is mainly neutral and the overall emotion of comments is neutral with pessimistic emotion. However, it can be seen that the temporal emotion evolution is not as strong as Figure~\ref{fig6}. Thus, the classification result of this case is true. This case shows the ability of proposed model DIDA of classifying true news by recognizing the temporal emotion evolution pattern of comments. 

Therefore, these two cases further demonstrate the high performance of our proposed model DIDA.

\section{Conclusion}
We propose a data-enhanced dual deep interaction channel network for fake news detection. In particular, we model the chronological dynamics of news-related comment emotion to uncover unique patterns of fake news that differ from real news. We design an interactive co-attention mechanism for news and comments in the emotion and semantic dual channels to fully interact with the representations. In addition, we fully utilize and enrich the annotated data by confidence threshold screening. The experimental results demonstrate the effectiveness of the method.

\bibliographystyle{ieeetr}
\bibliography{B}

\newpage

\section{Biography Section}

\vspace{11pt}

\begin{IEEEbiography} [{\includegraphics[width=1in,height=1.25in,clip,keepaspectratio]{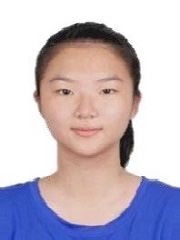}}]{Biwei Cao}
received the B.S. degree in Software Engineering from Australian National University, in 2019, the M.S. degree in Computing from Australian National University, in 2020. She is now working toward the Ph.D degree from the School of Cyber Science and Engineering, Southeast, University, Nanjing, China, Her research interests include affective computing, social computing and natural language processing.
\end{IEEEbiography}

\begin{IEEEbiography} [{\includegraphics[width=1in,height=1.25in,clip,keepaspectratio]{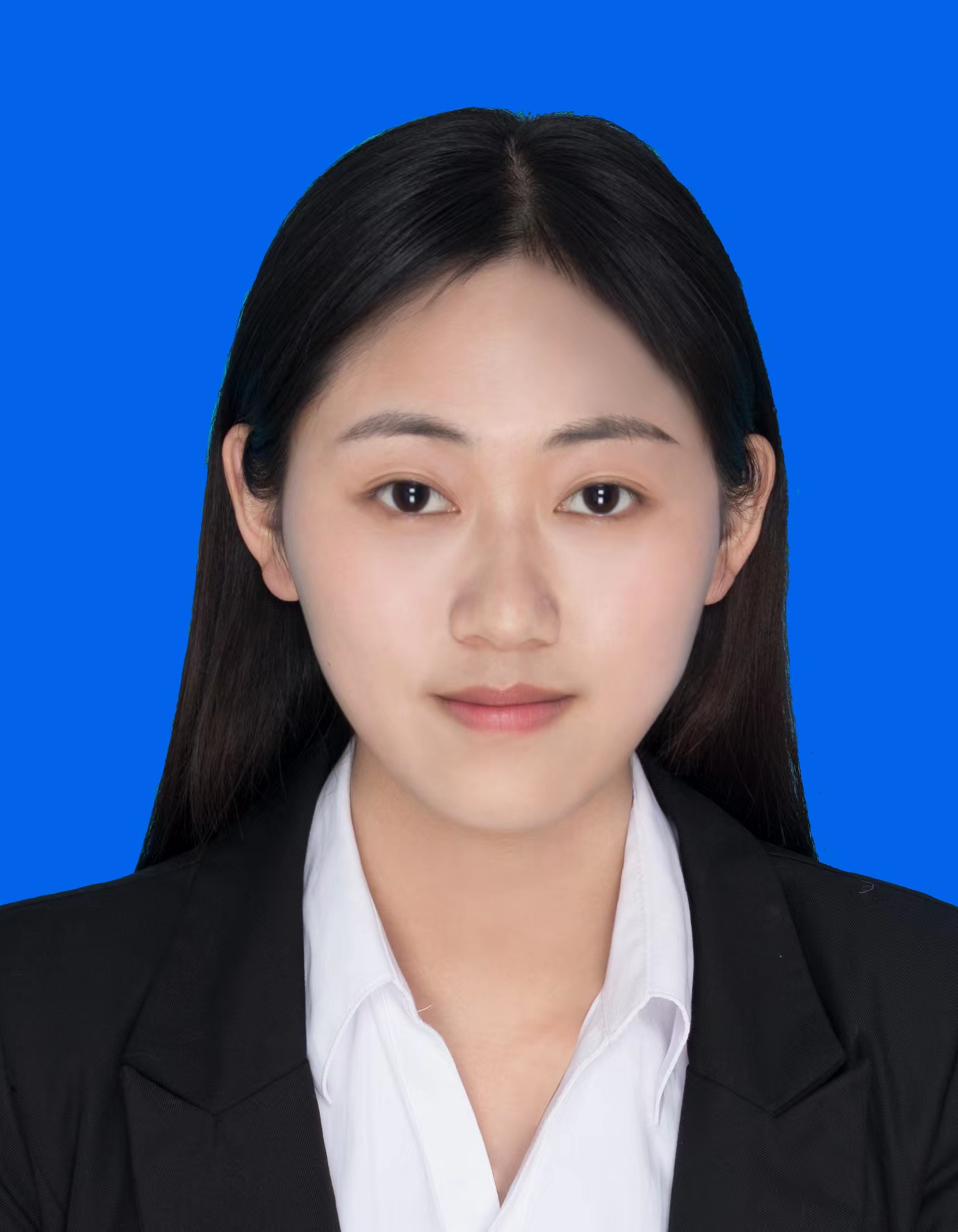}}]{Jiayun Shen} received the M.S. degree from the School of Cyber Science and Engineering, Southeast University, in 2020. His research interests includes social computing and natural language processing.
\end{IEEEbiography}

\begin{IEEEbiography} [{\includegraphics[width=1in,height=1.25in,clip,keepaspectratio]{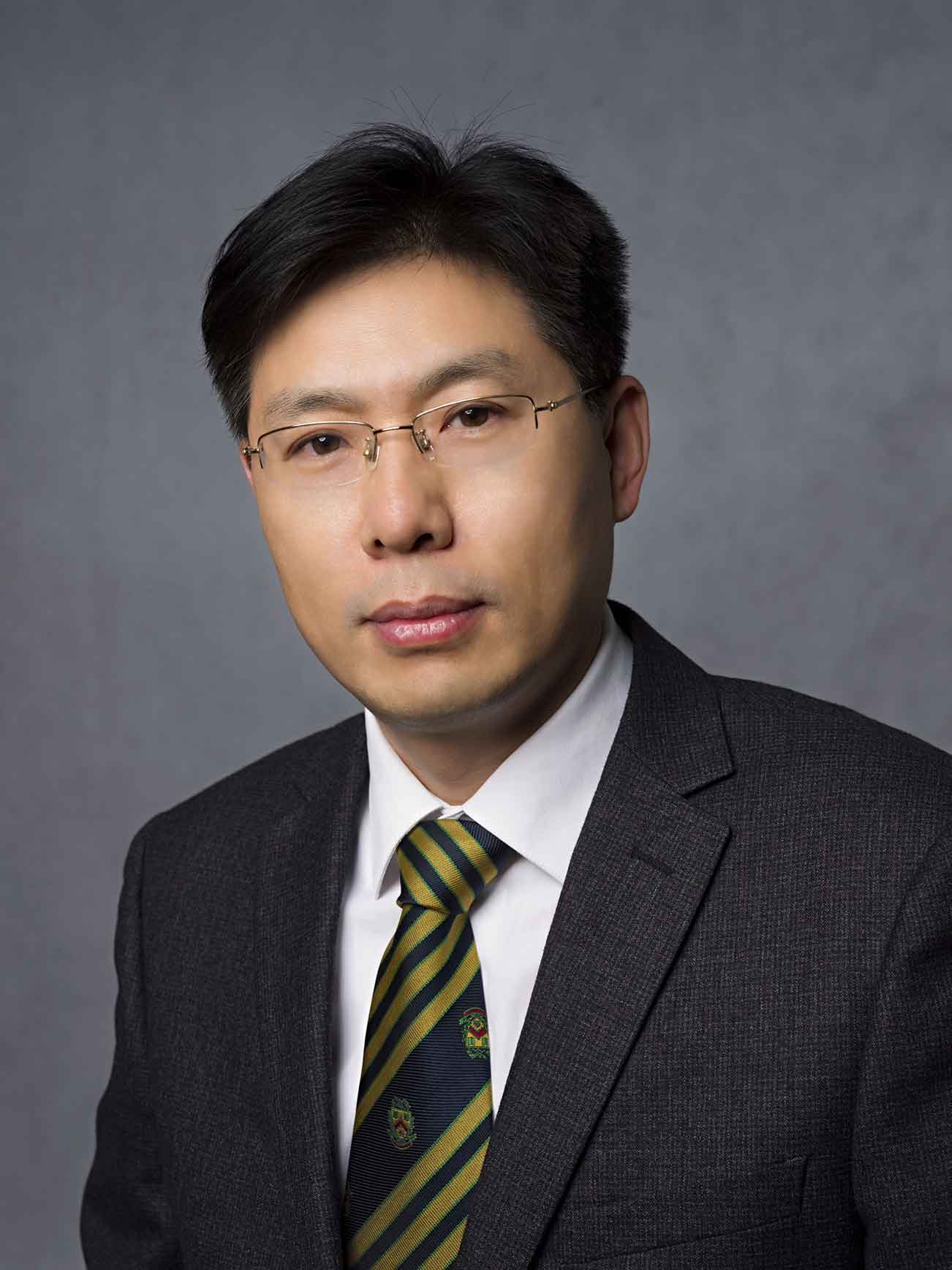}}]{Jiuxin Cao}
received the Ph.D degree from Xi’an Jiaotong University in 2003. He is currently a professor at the School of Cyber Science and Engineering in Southeast University, Nanjing, China. His research interests include computer networks, social computing, affective computing, behavior analysis, and big data security and privacy preservation.
He is the Director of Jiangsu Provincial Key Laboratory of Computer Network Technology, the Senior Member of China Computer Federation, the Member of Chinese Information Processing Society of China, the Fellow of Jiangsu Computer Society, the Member of Jiangsu Information Security Standardization Committee, and the Member of JSAI-ISA.
\end{IEEEbiography}

\begin{IEEEbiography} [{\includegraphics[width=1in,height=1.25in,clip,keepaspectratio]{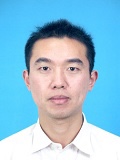}}]{Jie Gui}
 (Senior Member, IEEE) received the M.S. degree in computer applied technology from Hefei Institutes of Physical Science, Chinese Academy of Sciences, Hefei, China, in 2007, and the Ph.D. degree in pattern recognition and intelligent systems from the University of Science and Technology of China, Hefei, China, in 2010. He is currently a Professor in the School of Cyber Science and Engineering, Southeast University. He has published more than 40 papers in international journals and conferences such as IEEE TPAMI, IEEE TNNLS, IEEE TCYB, IEEE TIP, IEEE TCSVT, IEEE TSMCS, KDD, and ACM MM. He is the Area Chair, Senior PC member, or PC Member of many conferences such as NeurIPS and ICML. His research interests include machine learning, pattern recognition, and image processing.
\end{IEEEbiography}

\begin{IEEEbiography} [{\includegraphics[width=1in,height=1.25in,clip,keepaspectratio]{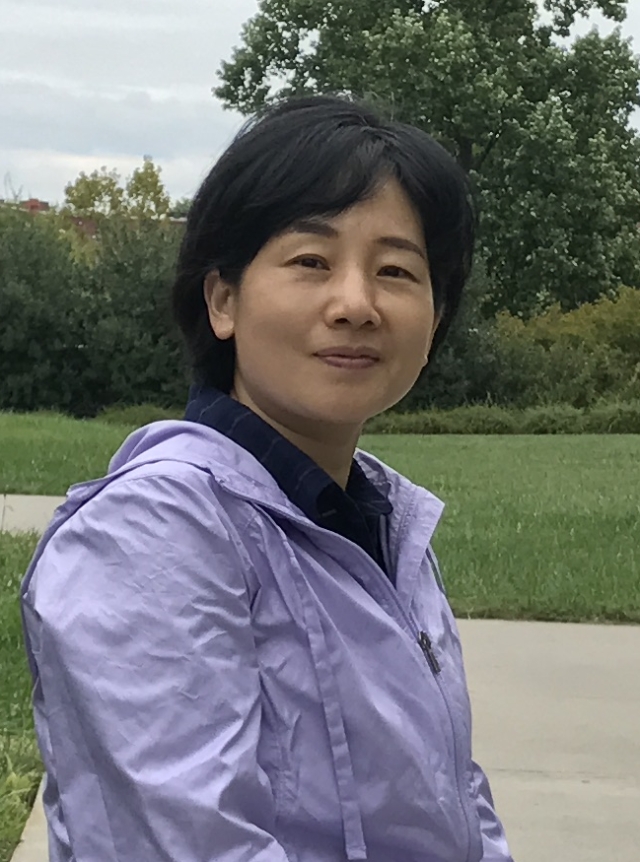}}]{Bo Liu}
(Member, IEEE) works as a professor and the doctoral advisor with Southeast University, China. She received her
doctoral degree from Southeast University. She won the first class Science and Technology Progress Award of MoE in 2009, and she is currently working on two NSF projects. She has published more than 60 papers and most of them have been published in reputed journals and conferences including WWW, WWWJ, ToN and et al. Her current main research interests include spammer detection in social network, the evolution of social community, social influence, and social recommendation.
\end{IEEEbiography}

\begin{IEEEbiography} [{\includegraphics[width=1in,height=1.25in,clip,keepaspectratio]{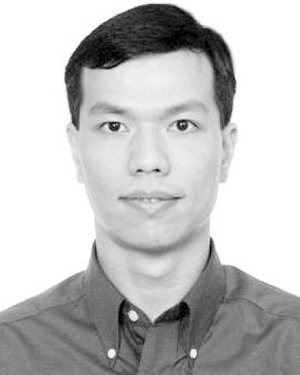}}]{James Tin-Yau Kwok}
(Fellow, IEEE) received the Ph.D. degree in computer science from The Hong Kong University of Science and Technology, Hong Kong, in 1996. He is currently a Professor with the Department of Computer Science and Engineering, The Hong Kong University of Science and Technology. His current research interests include kernel methods, machine learning, pattern recognition, and artiﬁcial neural networks. He received the IEEE Outstanding Paper Award in 2004 and the Second Class Award in Natural Sciences from the Ministry of Education, China, in 2008. He has been a Program Co-Chair for a number of international conferences, and served as an Associate Editor for the IEEE TRANS-ACTIONS ON NEURAL NETWORKS AND LEARNING SYSTEMS from 2006 to 2012. He is currently an Associate Editor of \textit{Neurocomputing}.
\end{IEEEbiography}
\vfill

\end{document}